%% file: main.tex
\documentclass[final]{cvpr}

\usepackage{times}
\usepackage{epsfig}
\usepackage{graphicx}
\usepackage{amsmath}
\usepackage{amssymb}
\usepackage{enumitem}
\usepackage{multirow}
\usepackage{pifont}
\usepackage{amstext}
\usepackage{float}
\setlist[itemize]{leftmargin=*}
\setlist[enumerate]{leftmargin=*}

\usepackage[pagebackref=true,breaklinks=true,colorlinks,bookmarks=false]{hyperref}

\def\cvprPaperID{2715} 
\def\confYear{CVPR 2021}
\usepackage{caption} 

\begin{document}

\title{Function4D: Real-time Human Volumetric Capture from\\ Very Sparse Consumer RGBD Sensors}


\author{
	Tao Yu\textsuperscript{1},
	Zerong Zheng\textsuperscript{1},
	Kaiwen Guo\textsuperscript{2},
	Pengpeng Liu\textsuperscript{3},
	Qionghai Dai\textsuperscript{1},
	Yebin Liu\textsuperscript{1}
	\\
	\textsuperscript{1}Department of Automation, Tsinghua University, Beijing, China
	\quad
	\textsuperscript{2}Google, Switzerland
	\\
	\textsuperscript{3}Institute of Automation, Chinese Academy of Sciences, Beijing, China
}

\maketitle

\begin{abstract}
Human volumetric capture is a long-standing topic in computer vision and computer graphics. 
Although high-quality results can be achieved using sophisticated off-line systems, real-time human volumetric capture of complex scenarios, especially using light-weight setups, remains challenging. 
In this paper, we propose a human volumetric capture method that combines temporal volumetric fusion and deep implicit functions.  
To achieve high-quality and temporal-continuous reconstruction, we propose dynamic sliding fusion to fuse neighboring depth observations together with topology consistency. 
Moreover, for detailed and complete surface generation, we propose detail-preserving deep implicit functions for RGBD input which can not only preserve the geometric details on the depth inputs but also generate more plausible texturing results. 
Results and experiments show that our method outperforms existing methods in terms of view sparsity, generalization capacity, reconstruction quality, and run-time efficiency. 
\end{abstract}

\input{1_intro}
\input{2_related_work}

\input{3_overview}
\input{4_fusion}
\input{5_infer}

\input{6_results}
\input{7_conclusion}

\clearpage
{\small
\bibliographystyle{ieee_fullname}
\bibliography{egbib}
}

\end{document}

%% file: 1_intro.tex
\section{Introduction}
\label{sec:intro}
Real-time volumetric capture of human-centric scenarios is the key to a large number of applications ranging from telecommunications, education, entertainment, and so on. 
And the underlying technique, volumetric capture, is a challenging and long-standing problem in both computer vision and computer graphics due to the complex shapes, fast motions, and changing topologies (e.g., human-object manipulations and multi-person interactions) that need to be faithfully reconstructed. 
Although high-end volumetric capture systems \cite{bradley2008markerless, gall2009motion, liu2009point, brox2009combined, liu2011markerless, pons2017clothcap} based on dense camera rigs (up to 100 cameras \cite{collet2015high}) and custom-designed lighting conditions \cite{VlasicPBDPRM09,guo2019relightables} can achieve high-quality reconstruction, they all suffer from complicated system setups and are limited to professional studio usage. 

In contrast, light-weight volumetric/performance capture systems are more practical and attractive. 
Given a pre-scanned template, ~\cite{li2009robust, zollhofer2014real, guo2015robust} track dense surface deformations from single-view RGB input~\cite{habermann2019livecap,habermann2020deepcap}. However, the prerequisite of a fixed-topology template restricts their applications for general volumetric capture. 
In 2015, DynamicFusion~\cite{newcombe2015dynamicfusion} proposed the first template-free and single-view dynamic 3D reconstruction system. The following works ~\cite{yu2017BodyFusion, yu2018doublefusion, Xu2020UnstructuredFusion, su2020robustfusion}
further improve the reconstruction quality for human performance capture by
incorporating semantic body priors. However, it remains challenging for them to handle large topological changes like dressing or taking-off clothes. 
Recently, a line of research \cite{natsume2019siclope,saito2019pifu,saito2020pifuhd,li2020monoport} leverages deep implicit functions for textured 3D human reconstruction only from a single RGB image.
However, they still suffer from off-line reconstruction performance ~\cite{saito2020pifuhd,saito2019pifu} or over-smoothed, temporally discontinuous results~\cite{li2020monoport}.
State-of-the-art real-time volumetric capture systems are volumetric fusion methods like Fusion4D~\cite{dou2016fusion4d} and Motion2Fusion~\cite{Motion2Fusion}. But both of them rely on custom-designed high-quality depth sensors (up to 120 fps and 1k resolution) and multiple (up to 9) high-end GPUs, which is infeasible for consumer usage.  

In this paper, we propose Function4D, a volumetric capture system using very sparse (as sparse as 3) consumer RGBD sensors. Compared with existing systems, our system is able to handle various challenging scenarios, including human-object manipulations, dressing or taking off clothes, fast motions and even multi-person interactions, as shown in Fig.~\ref{fig_teaser}. 


Our key observations are: 
To generate complete and temporal consistent results, current volumetric fusion methods have to fuse as much temporal depth observations as possible. This results in heavy dependency on accurate and long-term non-rigid tracking, which is especially challenging under severe topology changes and large occlusions. 
On the contrary, deep implicit functions are good at completing surfaces, but they cannot recover detailed and temporal continuous results due to the insufficient usage of depth information and severe noise from consumer RGBD sensors. 

To overcome all the limitations above, we propose a novel volumetric capture framework that organically combines volumetric fusion with deep implicit functions. 
By introducing dynamic sliding fusion, we re-design the volumetric fusion pipeline to restrict tracking and fusion in a sliding window and finally got noise-eliminated, topology-consistent, and temporally-continuous volumetric fusion results. 
Based on the sliding fusion results, we propose detail-preserving deep implicit functions for final surface reconstruction to eliminate the heavy dependency on long-term tracking. Moreover, by encoding truncated projective SDF (PSDF) values explicitly and incorporating attention mechanism into the multi-view feature aggregation stage, our networks not only achieve detailed reconstruction results but also orders of magnitude faster than existing methods. 

Our contributions can be summarized as:
\begin{itemize}
    \item The first real-time volumetric capture system which combines volumetric fusion with deep implicit functions using very sparse consumer RGBD sensors. 
    \item Dynamic Sliding Fusion for generating noise-eliminated and topology consistent volumetric fusion results.
    \item Detail-preserving Implicit Functions specifically designed for sufficient utilization of RGBD information to generate detailed reconstruction results. 
    \item The training and evaluation dataset, which contains 500 high-resolution scans of various poses and clothes, will be publicly available to stimulate future research. \end{itemize} 

%% file: 2_related_work.tex
\begin{figure*}
    \centering
    \includegraphics[width=0.98\linewidth]{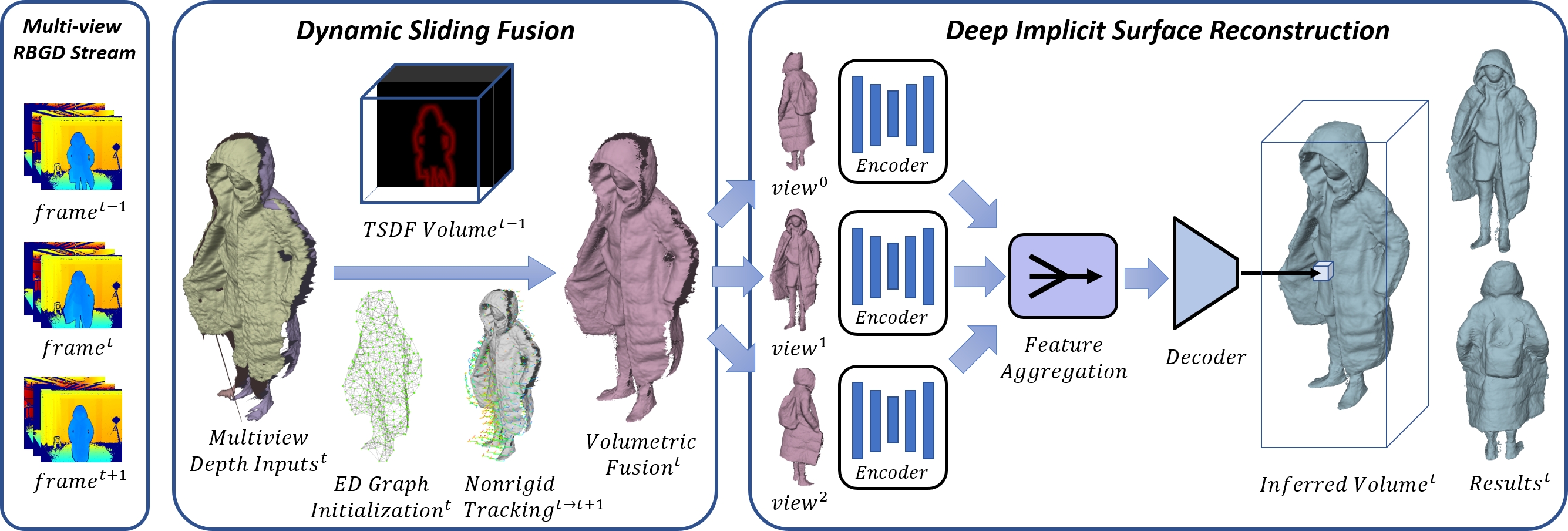}
    \caption{Overview. We focusing on geometry reconstruction in this figure. For color inference please refer to Fig.\ref{fig_network_structures}. }
    \label{fig_overview}
\end{figure*}

\section{Related Work}
\label{sec:related}
In the following, we focus on 3D human volumetric/performance capture and classify existing methods into $4$ categories according to their underlying techniques. 

\noindent\textbf{Volumetric capture from multi-view stereo. } 
Multi-view volumetric capture is an active research area in the computer vision and graphics community. Previous works use multi-view images for human model reconstruction~\cite{Kanade97,StarckCGA07,liu2009point}. Shape cues like silhouette, stereo, shading, and cloth priors have been integrated to improve reconstruction/rendering performance ~\cite{StarckCGA07,liu2009point,WuShadingHuman,WaschbuschWCSG05,VlasicPBDPRM09,bradley2008markerless,Mustafa16,pons2017clothcap,Wu_2020_CVPR}. 
State-of-the-art methods build extremely sophisticated systems where up to 100 cameras~\cite{collet2015high} and even custom-designed gradient lighting~\cite{guo2019relightables} for high-quality volumetric capture. In particular, the methods by~\cite{collet2015high} and~\cite{guo2019relightables} first perform multi-view stereo for the point cloud generation, followed by mesh construction, simplification, tracking, and post-processing steps such as UV mapping. Although the results are compelling, the reliance on well-controlled multi-camera studios and a huge amount of computational resources prohibit them from being used in living spaces. 

\noindent\textbf{Template-based Performance Capture. }
For performance capture, some of the previous works leverage pre-scanned templates and exploit multi-view geometry to track the motion of the templates. For instance, the methods in~\cite{Vlasic08,gall2009motion,brox2009combined} adopted a template with an embedded skeleton driven by multi-view silhouettes and temporal feature constraints. Their methods are then extended to handle multiple interacting characters in~\cite{liu2011markerless,LiuPAMI13}. Besides templates with an embedded skeleton, some works adopted non-rigid deformation for template motion tracking. Li \textit{et,al}~\cite{li2009robust} utilized embedded deformation graph in~\cite{Sumner2007embedded} to parameterize the non-rigid deformations of the pre-scanned template. Guo \textit{et,al.}~\cite{guo2015robust} adopted an $\ell_0$ norm constraint to generate articulated motions for bodies and faces without explicitly constructing a skeleton. Zollh{\"o}fer \textit{et, al.}~\cite{zollhofer2014real} took advantage of GPU to parallelize the non-rigid registration algorithm and achieved real-time performance of general non-rigid tracking. Recently, capturing 3D dense human body deformation with coarse-to-fine registration from a single RGB camera has been enabled~\cite{xu2018monoperfcap} and improved for real-time performance~\cite{habermann2019livecap}. DeepCap~\cite{habermann2020deepcap} introduced a deep learning method that jointly infers the articulated and non-rigid 3D deformation parameters in a single feed-forward pass. Although template-based approaches require less input than multi-view stereo methods, they are incapable of handling topological changes due to the prerequisite of a template with fixed topology. 


\noindent\textbf{Volumetric Fusion for Dynamic 3D Reconstruction. }
To get rid of template priors and realize convenient deployment, researchers turn to use one or sparse depth sensors for 3D reconstruction. In 2015, DynamicFusion~\cite{newcombe2015dynamicfusion} proposed the first template-free, single-view, real-time dynamic 3D reconstruction system, which integrates multiple frames into a canonical model to reconstruct a complete surface model. However, it only handles controlled and relatively slow motions due to the challenges of real-time non-rigid tracking. In order to improve the robustness of DynamicFusion, the following works incorporated killing/sobolev constraints~\cite{slavcheva2017killingfusion, Slavcheva_2018_CVPR}, articulated priors~\cite{chao2018ArticulatedFusion}, skeleton tracking~\cite{yu2017BodyFusion}, parametric body models~\cite{yu2018doublefusion}, sparse inertial measurements~\cite{Zheng2018HybridFusion}, data-driven prior~\cite{su2020robustfusion} or learned correspondences~\cite{bozic2020deepdeform} into the non-rigid fusion pipeline. However, all of these methods are prone to tracking failure in invisible areas, which is an inherent drawback of single-view systems. To overcome this limitation, Fusion4D~\cite{dou2016fusion4d} and Motion2fusion~\cite{Motion2Fusion} focused on real-time multi-view setups using high-end custom-design sensors, with the notion of key volume updating and learning-based surface matching. Even though the pipelines were carefully designed in~\cite{dou2016fusion4d} and~\cite{Motion2Fusion}, they still suffer from incomplete and noisy reconstructions when severe topological changes occur especially under very sparse system setups.

\noindent\textbf{Learning-based 3D Human Reconstruction. }
Fueled by the rapid developments in neural 3D representations(e.g., ~\cite{park2019deepsdf,Peng2020ECCV,DeepLocalShapes,Occupancy_Networks} etc.), a lot of data-driven methods for 3D human reconstruction have been proposed in recent years. 
Methods in~\cite{zhu2019hmd,alldieck2019tex2shape} proposed to deform a parametric body model to fit the image observations including keypoints, silhouettes, and shading.  DeepHuman~\cite{zheng2019deephuman} combined the parametric body model with a coarse-scale volumetric reconstruction network to reconstruct 3D human models from a single RGB image. Some methods infer human shapes on 2D image domains using multi-view silhouettes~\cite{natsume2019siclope} or front-back depth pairs~\cite{gabeur2019moulding,wang2020normalgan}. PIFu~\cite{saito2019pifu} proposed to regress an implicit function using pixel-aligned image features. Unlike voxel-based methods, PIFu is able to reconstruct high-resolution results thanks to the compactness of the implicit representations. PIFuHD~\cite{saito2020pifuhd} extended PIFu to capture more local details. However, both PIFu~\cite{saito2019pifu} and PIFuHD~\cite{saito2020pifuhd} fail to reconstruct plausible models in cases of challenging poses and self-occlusions. PaMIR~\cite{zheng2020pamir} resolved this challenge by using the SMPL model as a prior but suffers from run-time inefficiency since it requires a post-processing optimization step. 
IFNet~\cite{chibane20ifnet} and IPNet~\cite{bhatnagar2020ipnet} can recover impressive 3D humans from partial point clouds, but the dependency on multi-scale 3D convolutions and parametric body models block the realization of real-time reconstruction performance.

%% file: 3_overview.tex
\section{Overview}
\label{sec:overview}
As shown in Fig.~\ref{fig_overview}, the proposed volumetric capture pipeline mainly contains 2 steps: Dynamic Sliding Fusion and Deep Implicit Surface Reconstruction. 
Given a group of synchronized multi-view RGBD inputs, we first perform dynamic sliding fusion by fusing its neighboring frames to generate noise-eliminated and temporal-continuous fusion results. 
After that, we re-render multi-view RGBD images using the sliding fusion results in the original viewpoints. 
Finally, in the deep implicit surface reconstruction step, we propose detail-preserving implicit functions (which consists of multi-view image encoders, a feature aggregation module, and an SDF/RGB decoder) for generating detailed and complete reconstruction results. 

%% file: 4_fusion.tex
\begin{figure}[ht]
    \centering
    \includegraphics[width=0.9\linewidth]{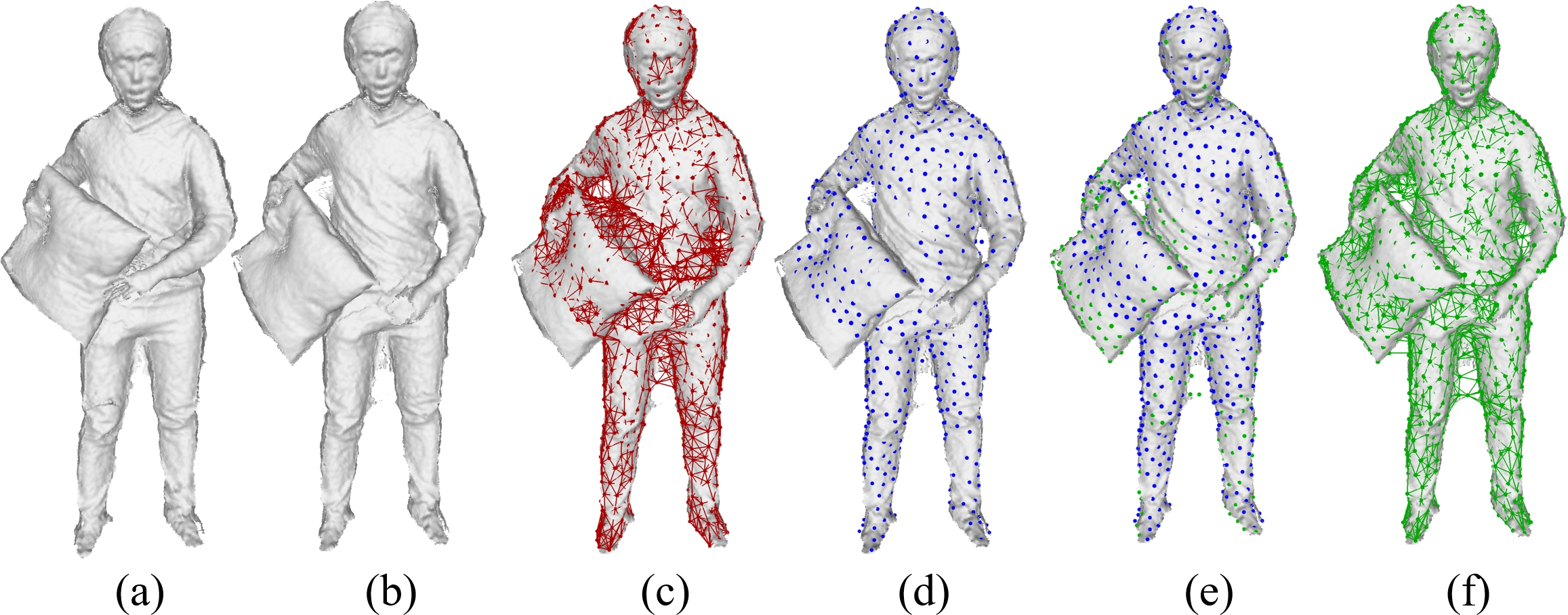}
    \caption{Topology-aware node graph initialization. (a) and (b) are the preceding and the current frame, respectively. (c) is the overlay of the current frame and the warped node graph of the preceding frame; erroneous nodes can be found around the pillow and arms where large topological changes occur. (d), (e) and (f) are the cleared node graph, the additional nodes (green) for newly observed surfaces, and the final node graph for the current frame, respectively. }
    \label{fig_node_graph}
\end{figure}

\section{Dynamic Sliding Fusion (DSF)}

Different from previous volumetric fusion methods, the proposed DSF method aims at augmenting current observations but not completing surfaces. So we re-design the fusion pipeline to make sure we can get topologically consistent and noise-eliminated results for current observations. 

The proposed DSF mainly contains 3 steps: topology-aware node graph initialization, non-rigid surface tracking, and observation-consistent truncated SDF (TSDF) fusion. To eliminate the heavy dependency on long-term tracking, instead of fusing all frames as in previous methods, we only perform fusion in a sliding window in DSF.
Specifically, we allow a 1-frame delay for the reconstruction pipeline and fuse the current frame (indexed by t) only with its preceding frame (t-1) as well as its succeeding frame (t+1) to minimize the topological changes and tracking error accumulation. 
Note that we only need to perform non-rigid surface tracking for the succeeding frame since the deformation between the current frame and the preceding frame has already been tracked. 
Regarding the TSDF fusion stage, we propose observation-consistent TSDF fusion to fuse multi-view observations of frame t+1 into frame t. 



        
\subsection{Topology-aware Node Graph Initialization}
\label{sec:DSF_node_graph}
Previous fusion methods initialize the embedded deformation graph (ED graph) ~\cite{Sumner2007embedded} in the canonical frame~\cite{newcombe2015dynamicfusion}. However, such an ED graph cannot well describe the topological changes in live frames. Different from previous methods, we have to initialize the ED graph exactly for the current frame to guarantee that the topology of the node graph is consistent with the current observations. 
However, it is inefficient to initialize the node graphs from scratch for every frame because of the complexity of node selection, graph connection, and volume KNN-field calculation. 
To overcome this limitation, we propose topology-aware node graph initialization, which can not only leverage the node graph of the previous frame for fast initialization but also has the ability to generate a topologically-consistent node graph for the current observations. 

As shown in Fig.~\ref{fig_node_graph}, we first initialize the current node graph using the live node graph from the preceding frame. This is achieved by warping the node graph of the preceding frame to the current frame directly.
Due to tracking errors and topological changes, the live node graph may not be well aligned with current observations (Fig.~\ref{fig_node_graph}(c)). So we clear those nodes that are located far from current observations by constraining their TSDF values in the current TSDF volume (Fig.~\ref{fig_node_graph}(d)). Specifically, if the magnitude of the normalized TSDF value corresponding to a node is greater than $\delta_t$, we suppose that it is relatively far from the observations and we delete this node to maintain the current mesh topology.
Finally, considering that there may still exist newly observed surfaces that are not covered by the cleared node graph, we further refine it based on the current observations by sampling additional nodes (Fig.~\ref{fig_node_graph}(e)) as in DynamicFusion~\cite{newcombe2015dynamicfusion} to make sure that all of the surfaces can be covered by the final node graph (Fig.~\ref{fig_node_graph}(f)). 
    
\subsection{Non-rigid Surface Tracking}
\label{sec:DSF_nonrigid}
For non-rigid surface registration, we follow previous methods to search projective point-to-plane correspondences between the surface at frame t and the multi-view depth observations at frame t+1. 
The definition of the non-rigid tracking energy is: 
\begin{equation} \label{eqn:tracking_energy}
E_{\mathrm{tracking}} = \lambda_{\mathrm{data}}E_{\mathrm{data}} + \lambda_{\mathrm{reg}}E_{\mathrm{reg}},
\end{equation}
where $E_{\mathrm{data}}$ and $E_{\mathrm{reg}}$ are the energies of data term and regularization term respectively. The data term measures the fitting error between the deformed surface and the depth observations, while the regularization term enforces local as-rigid-as-possible surface deformations. 
A Gauss-Newton solver with Preconditioned Conjugate Gradient algorithm (PCG) is used to solve this non-linear optimization problem efficiently on GPU. Please refer to ~\cite{newcombe2015dynamicfusion} for more details. 

\begin{figure}[]
    \centering
    \includegraphics[width=0.9\linewidth]{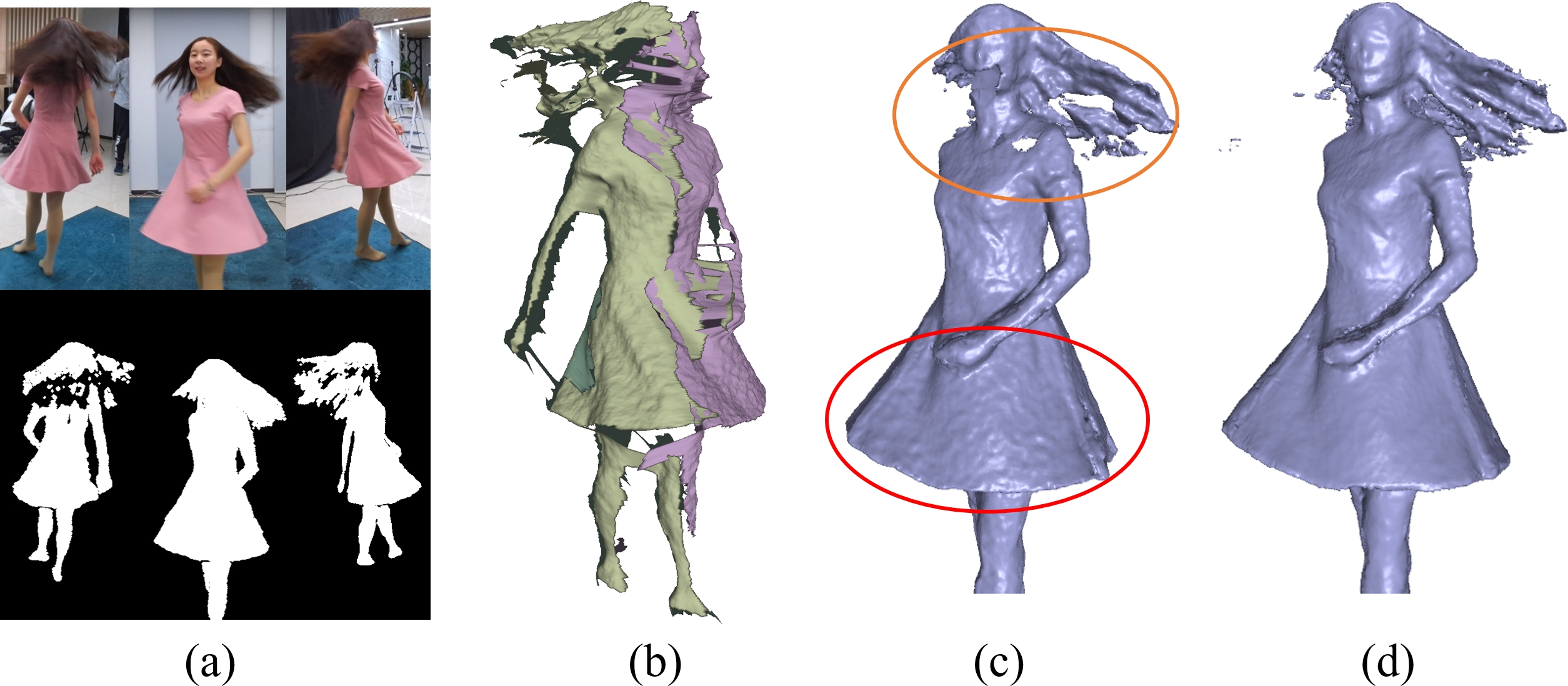}
    \caption{Evaluation of dynamic sliding fusion. From (a) to (d) are multi-view RGB references and depth masks, multi-view depth input rendered in a side viewpoint, results without and with dynamic sliding fusion, respectively. }
    \label{fig_eval_fusion}
\end{figure}

\subsection{Observation-consistent TSDF Fusion}
\label{sec:DSF_fusion} 
After non-rigid surface tracking, we warp the volume of frame t to the depth observations at frame t+1 for TSDF fusion. Since we are focusing on augmenting the current observations but not completing surfaces, we propose an aggressive TSDF fusion strategy to eliminate the impact of tracking errors and severe topological changes. 
Specifically, in our fusion method, a voxel in the volume of frame t will update its TSDF value if and only if: i) the tracking error corresponding to this voxel is lower than a threshold $\delta_e$, and ii) there exist valid voxels (voxels that in the truncated band of current observations) located around this voxel (the searching radius is set to $3$ in our implementation). 

Calculating the tracking error for a specific voxel is not straightforward since non-rigid tracking is established between the reference surface and the depth maps~\cite{Motion2Fusion}. So we first calculate the tracking error for each node on the ED graph and then calculate the tracking error for each voxel by interpolation. 
Suppose $\mathcal{C}={(v_i,p_i)}$ is the correspondence set after the last iteration of non-rigid tracking, where $v_i$ is the $i$th vertex on the reference surface and $p_i$ is the corresponding point of $v_i$ on live depth inputs. 
The tracking error corresponding to the $j$th node can be calculated as: 
\begin{equation} \label{eqn:node_tracking_error}
\mathbf{e}(n_j) = \frac{\sum_{v_i\in\mathcal{C}_j}\mathbf{r}(v_i',p_i)}{\sum_{v_i\in\mathcal{C}_j}\mathbf{w}(v_i,n_j) + \epsilon},
\end{equation}
where $\mathbf{r}(v_i',p_i) = \|pn_i^T\cdot(v_i'-p_i)\|_2^2$ is the residue of $v_i$ after non-rigid tracking, $v_i'$ the warped position of $v_i$, $pn_i$ the surface normal of $p_i$, $\mathcal{C}_j$ a subset of $\mathcal{C}$ which includes all the reference vertices controlled by node $j$, $\mathbf{w}(v_i,n_j)=exp(-\|v_i-n_j\|_2^2/(2d^2))$ the blending weight of node $j$ on $v_i$, in which $d$ is the influence radius of $n_j$ and $\epsilon=1e-6$ is used to avoid zero division. 

For a voxel $x_k$, its tracking error is then interpolated using its $K$-Nearest-Neighbors on the node graph $\mathcal{N}(x_k)$ (where $K=4$) as: 

\begin{equation} \label{eqn:node_tracking_error}
\mathbf{e}(x_k) = \sum_{j\in\mathcal{N}(x_k)}\mathbf{w}(x_k,n_j)\cdot\mathbf{e}(n_j).
\end{equation}

%% file: 5_Infer.tex
\section{Deep Implicit Surface Reconstruction}
\label{sec_DISR}
After dynamic sliding fusion, we can get noise-eliminated surfaces. However, the surfaces are by no means complete due to the very sparse inputs and occlusions. The goal of the deep implicit surface reconstruction step is to generate complete and detailed surface reconstruction results using deep implicit functions. 
Since we have already fused a 3D TSDF volume in dynamic sliding fusion, a straightforward idea is to use a 3D convolution-based encoder-decoder network to ``inpaint'' the volume. And the methods in ~\cite{chibane20ifnet,bhatnagar2020ipnet} have achieved complete 3D surface reconstruction results by proposing multi-scale 3D convolution networks. However, the dependency on inefficient 3D convolution limits their applications in real-time systems, and the huge memory consumption restricts them from generating high-resolution results. 
In contrast, real-time implicit surface reconstruction can be achieved using 2D pixel-aligned local features combined with positional encoding as shown in ~\cite{li2020monoport}. However, it was designed for using only RGB images and can only generate over-smoothed results. 
Finally, regarding the RGBD-based implicit functions proposed in ~\cite{li2020portrait}, we can see that simply adding depth input as an additional input channel still cannot preserve the geometric details on the depth inputs. 

To resolve the limitations above, we propose a new deep implicit surface reconstruction method that is specifically designed for RGBD input. 
The implicit surface reconstruction contains two steps: 
First, we re-render the multi-view RGBD images from the fused surface after dynamic sliding fusion. 
And then, given multi-view RGBD inputs, we propose detail-preserving implicit functions to reconstruct a complete surface with texture for the current frame. 


\subsection{Multi-view RGBD Re-rendering}
In this step, we re-render multi-view RGBD images from the fused surfaces using input camera viewpoints. 
The re-rendered RGBD images contain much less noise than the original inputs thanks to the dynamic sliding fusion step. 
Note that another benefit of multi-view RGBD re-rendering is that we can manually fix the perspective projection parameters for all the rendered RGBD images to make sure they are consistent with the projection parameters that were used for rendering the training dataset. 
 



\begin{figure}[t]
    \centering
    \includegraphics[width=\linewidth]{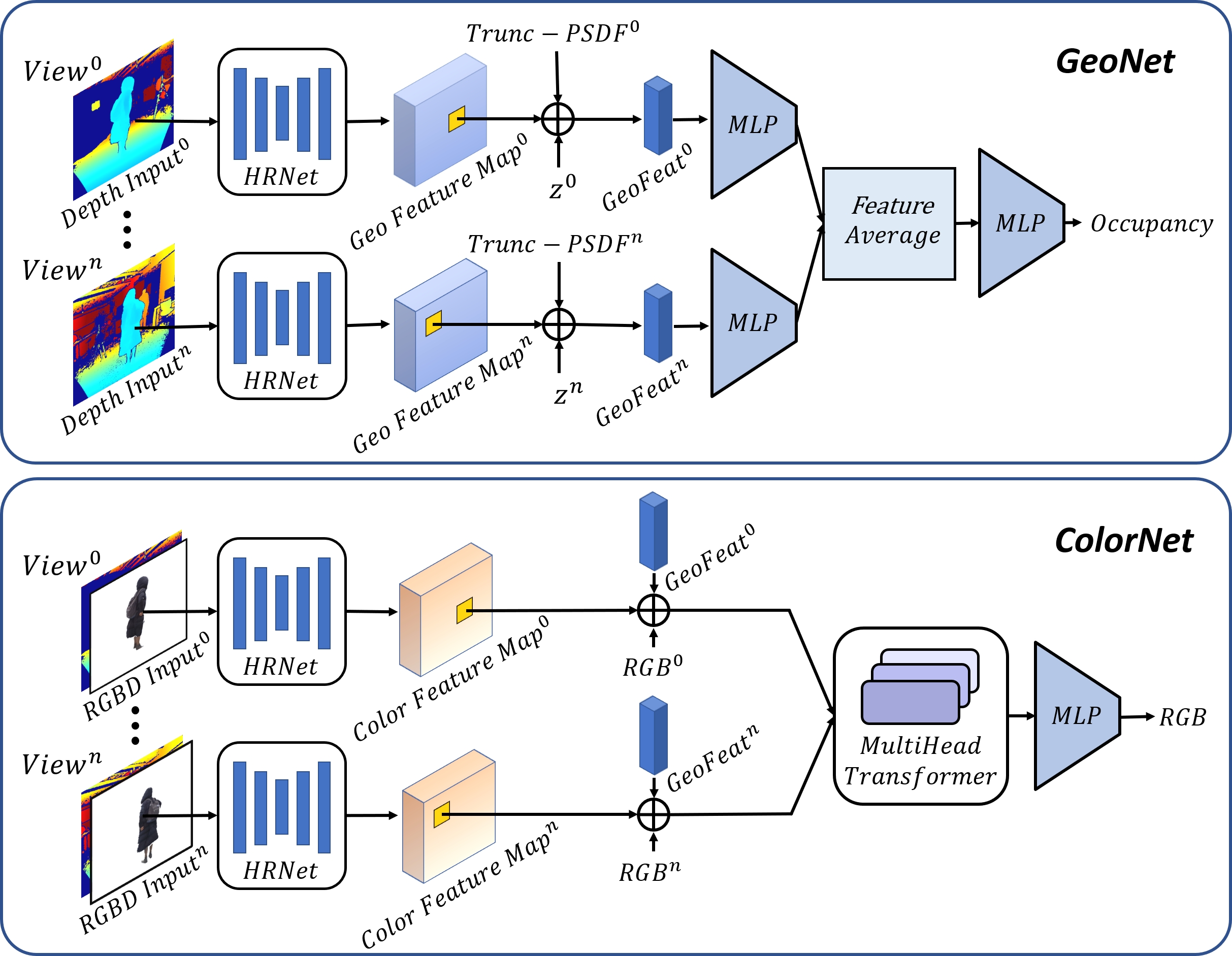}
    \caption{Network structures of GeoNet and ColorNet. }
    \label{fig_network_structures}
\end{figure}

\subsection{Detail-preserving Implicit Functions}
We propose two networks, \textit{GeoNet} and \textit{ColorNet}, for inferring detailed and complete geometry with color from multi-view RGBD images. 
As shown in Fig.~\ref{fig_network_structures}, GeoNet and ColorNet share similar network architectures. 
Different from~\cite{saito2019pifu,li2020monoport}, we explicitly calculate the truncated PSDF feature in \textit{GeoNet} for preserving geometric details on depth maps. Moreover, we use a multi-head transformer network for multi-view feature aggregation in \textit{ColorNet} to generate more plausible color inference results. 
Empirically, we found that using only depth images is enough for training the \textit{GeoNet}, so we eliminate the RGB information for geometry reconstruction for efficient reconstruction. 

\begin{figure}[t]
    \centering
    \includegraphics[width=0.9\linewidth]{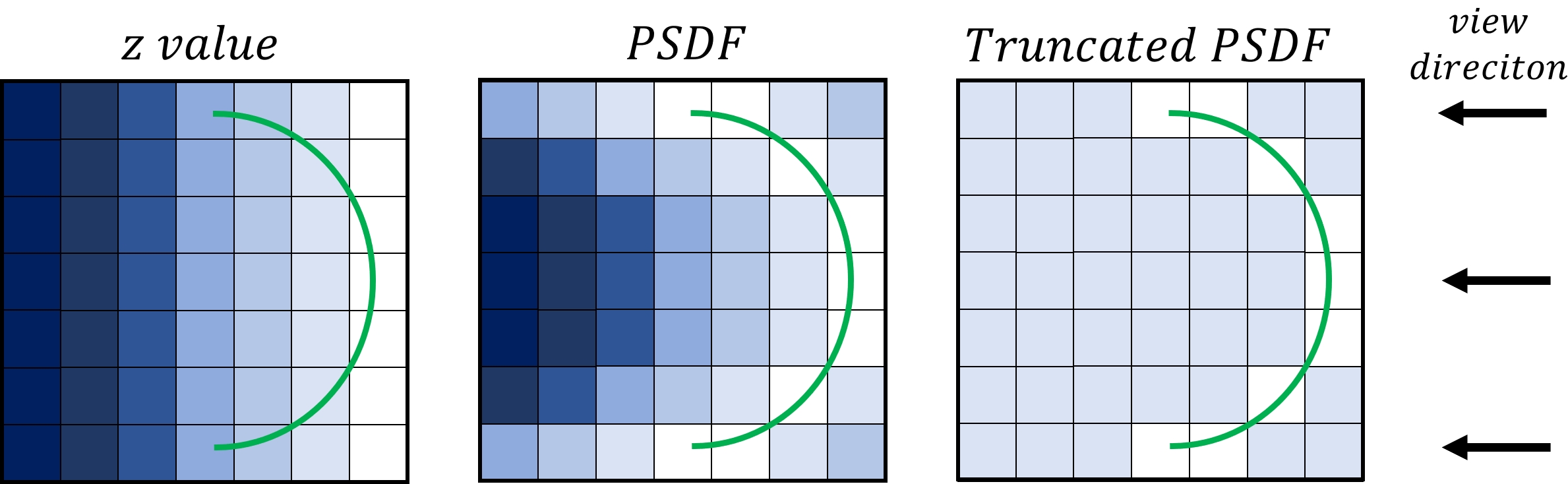}
    \caption{Illustration of the truncated PSDF feature. The green line represents the depth input. Note that we visualize the absolute of PSDF values here for simplicity, and the darker the grid, the larger the absolute PSDF values. 
    }
    \label{fig_illu_trunc_PSDF}
\end{figure}

\begin{figure}[t]
    \centering
    \includegraphics[width=0.9\linewidth]{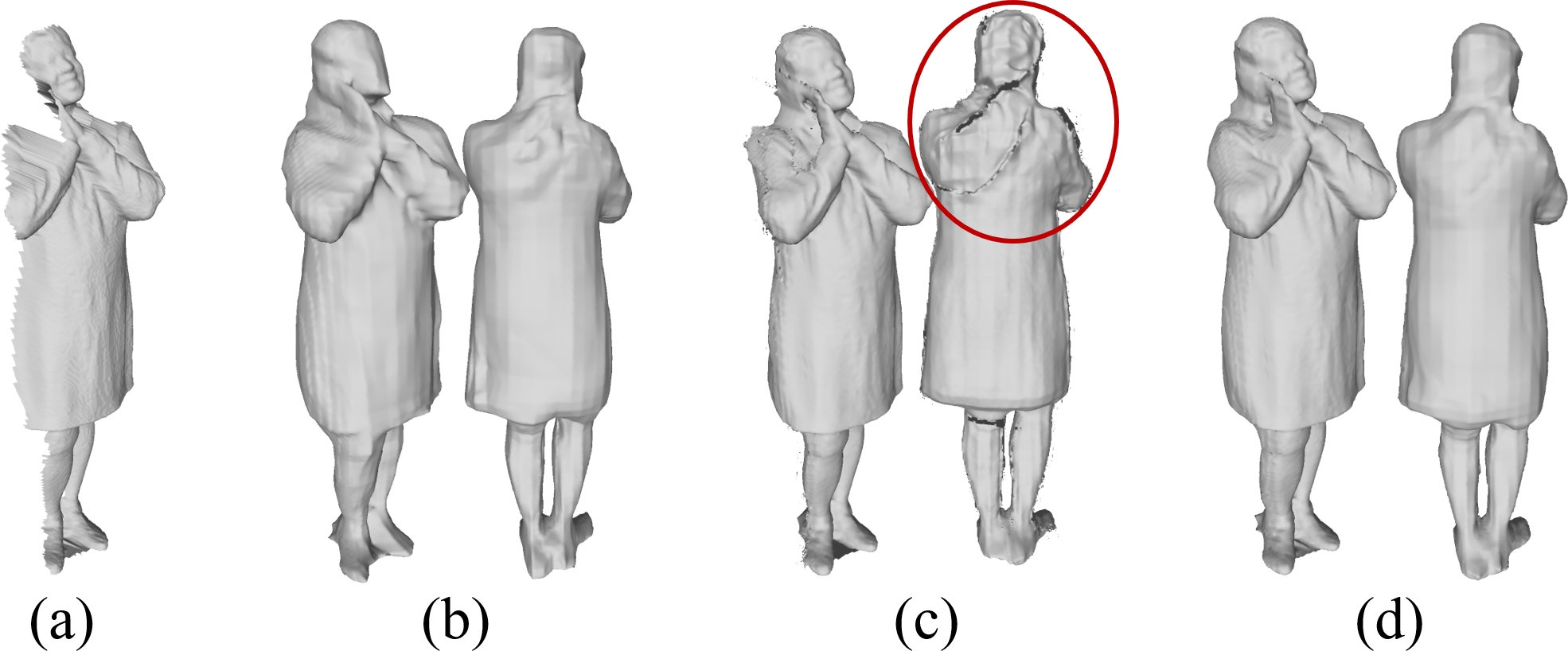}
    \caption{Evaluation of the truncated PSDF. (a) is single-view depth input, (b,c,d) are results generated without PSDF, with PSDF and with truncated PSDF, respectively. }
    \label{fig_eval_trunc_PSDF}
\end{figure}


\subsubsection{Truncated PSDF Feature}
The feature used for decoding occupancy values in pixel-aligned implicit functions can be decomposed into a 2D image feature and a positional encoding (z value in~\cite{saito2019pifu} or the one-hot mapping in~\cite{li2020monoport}). 
The previous method~\cite{li2020portrait} augments the 2D image feature by enhancing the 2D image feature using RGBD images as input. Although this can successfully guide the network to resolve the z-ambiguity of using only RGB images, it does not preserves the geometric details on the depth inputs. This is due to the fact that the variation of the geometric details on depth inputs is too subtle (when compared with the global range of depth inputs) for the networks to ``sense'' by 2D convolutions. 
To fully utilize the depth information, we propose to use truncated PSDF values as an additional feature dimension. 
The truncated PSDF value is calculated by: 
\begin{equation} \label{eqn:node_tracking_error}
\mathit{f}(q) = \mathbf{T}(q_{.z}-\mathbf{D}(\mathbf{\Pi}(q))),
\end{equation}
where $q$ is the coordinate of the query point, $\mathbf{\Pi}(\cdot)$ is the perspective projection function, $\mathbf{D}(\cdot)$ is a bi-linear sampling function used for fetching depth values on the depth image, 
and $T(\cdot)$ is used to truncate the PSDF values in $[-\delta_p,\delta_p]$. 

As shown in Fig.~\ref{fig_illu_trunc_PSDF}, the truncated PSDF value is a strong signal corresponding to the observed depth inputs. Moreover, it also eliminates the ambiguities of using global depth values. 
Fig.~\ref{fig_eval_trunc_PSDF}(b) demonstrates that without using the PSDF values, we can only get over-smoothed results even for the visible regions with detailed observations. Moreover, without truncation, the depth variations of the visible regions (the arms on top of the body) will be misleadingly transferred to the invisible regions (Fig.~\ref{fig_illu_trunc_PSDF}) and finally leads to ghost artifacts (the ghost arm in the red circle of Fig.~\ref{fig_eval_trunc_PSDF}(c)). 

\subsubsection{Multi-view Feature Aggregation}
Although PIFu~\cite{saito2019pifu} has demonstrated multi-view reconstruction results by averaging the intermediate feature of the SDF/RGB decoders, we argue that the average pooling operation has limited capacity and cannot capture the difference in inference confidence between different viewpoints. For color inference, the network should have the ability to ``sense'' the geometric structure and also the visibility in different viewpoints for a query point.

To fulfill this goal, we propose to leverage the attention mechanism in~\cite{Vaswani_attention_nips17} for multi-view feature aggregation in \textit{ColorNet}. 
Compared with direct averaging, the attention mechanism has the advantage of incorporating the inter-feature correlations between different viewpoints, which is necessary for multi-view feature aggregation. 
Intuitively, for a query point that is visible in $view^0$ but fully occluded in other views, the feature from the $view^0$ should play a leading role in the final decoding stage. 
As shown in Fig.~\ref{fig_eval_attn}, direct averaging of the multi-view features may lead to erroneous texturing results. On the contrary, using attention-based feature aggregation can enable more effective feature merging and thus generate more plausible color inference results. 
In practice, we follow~\cite{Vaswani_attention_nips17} to use multi-head self-attention with 8 heads and 2 layers without positional encoding. The input is the concatenation of multi-view geometry features, color features, and RGB values as in Fig.~\ref{fig_network_structures}. And we fuse the output multi-head features through a two layers FC and weighted summation. Moreover, we found that the attention mechanism has limited improvement for geometry reconstruction since the visibility has been encoded by the truncated PSDF feature in \textit{GeoNet}. 


%% file: 6_results.tex
\begin{figure*}[ht]
    \centering
    \includegraphics[width=0.9\linewidth]{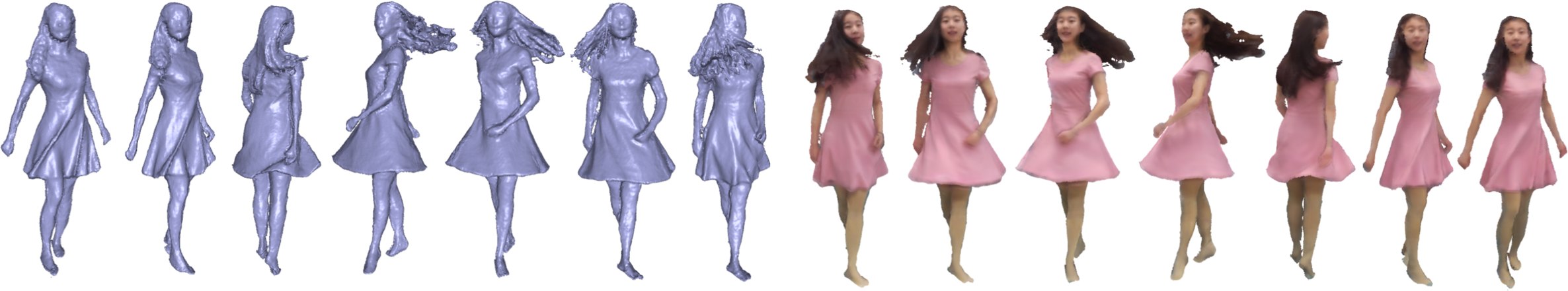}
    \caption{Temporal reconstruction results of a fast dancing girl. Our system can generate temporally-continuous and high-quality reconstruction results under challenging deformations (of the skirt) and severe topological changes (of the hair). }
    \vspace{-5pt}
    \label{fig_results}
\end{figure*}

\section{Results}
\label{sec:results}

The results of our system are shown in Fig.~\ref{fig_teaser} and Fig.~\ref{fig_results}. Note that the temporal-continuous results are reconstructed by our system under various challenging scenarios, including severe topological changes, human-object manipulations, and multi-person interactions. 

\subsection{Real-time Implementation}
In order to achieve real-time performance, we implement our run-time pipeline fully on GPU. 
Specifically, for deep implicit surface reconstruction, we use TensorRT with mixed precision for fast inference. 
After that, the major efficiency bottleneck of geometry inference lies in the evaluation of an excessive number of voxels when evaluating every voxel in the volume. 
Since we already have multi-view depth maps as input, we can leverage the depth information directly for acceleration without using the surface localization method in~\cite{li2020monoport}. 
Specifically, we first use the depth images to filter out empty voxels. Then we follow the octree-based reconstruction algorithm~\cite{Mescheder2019OccupancyNetwork} to perform inference for the rest voxels in a coarse-to-fine manner, which starts from a resolution of $64^3$ to the final resolution of $256^3$. 
To further improve the run-time efficiency, we simplify the network architectures as follows. For the image encoders, we follow~\cite{li2020monoport} and use HRNetV2-W18-Small-v2~\cite{SunXLW19Hrnet} as the backbone, setting its output resolution to $64\times 64$ and channel dimension to $32$. For the SDF/color decoders, we use MLPs with skip connections and the hidden neurons as $(128, 128, 128, 128, 128)$. 
For dynamic sliding fusion, we set $\delta_t=0.5$, $\delta_e=0.1$ and $\delta_p=0.01m$ for all the cases. We refer readers to~\cite{izadi2011kinectfusion,guo2017real} for real-time implementation details. For multi-view RGBD re-rendering, we render multi-view RGBD images in a single render pass with original color images as textures to improve efficiency. 
Finally, our system achieves reconstruction at $25fps$ with $21ms$ for dynamic sliding fusion, $17ms$ for deep implicit surface reconstruction (using $3$ viewpoints) and $2ms$ for surface extraction using Marching-Cubes Algorithm~\cite{lorensen1987marching}. 


\subsection{Network Training Details}
We use 500 high-quality scans for training \textit{GeoNet} and \textit{ColorNet}, which contains various poses, clothes and human-object interactions. 
We rotate each scan around the yaw axis, apply random shifts and render 60 views of the scan with image resolution of 512$\times$512. For color image rendering, we use the PRT-based rendering as in~\cite{saito2019pifu}. For depth rendering, we first render ground truth depth maps and then synthesis the sensor noises of TOF depth sensors on top of the depth maps according to~\cite{Kinectv2_noise_2015}. Note that we render all the RGBD images using perspective projection to keep consistent with real world sensors. During network training, gradient-based adaptive sampling (in which we use discrete gaussian curvature and rgb gradient as reference for query point sampling in \textit{GeoNet} and \textit{ColorNet} respectively) is used for more effective sampling around detailed regions. We randomly select 3 views from the rendered 60 views of a subject for multi-view training.

\begin{figure}[ht]
    \centering
    \includegraphics[width=\linewidth]{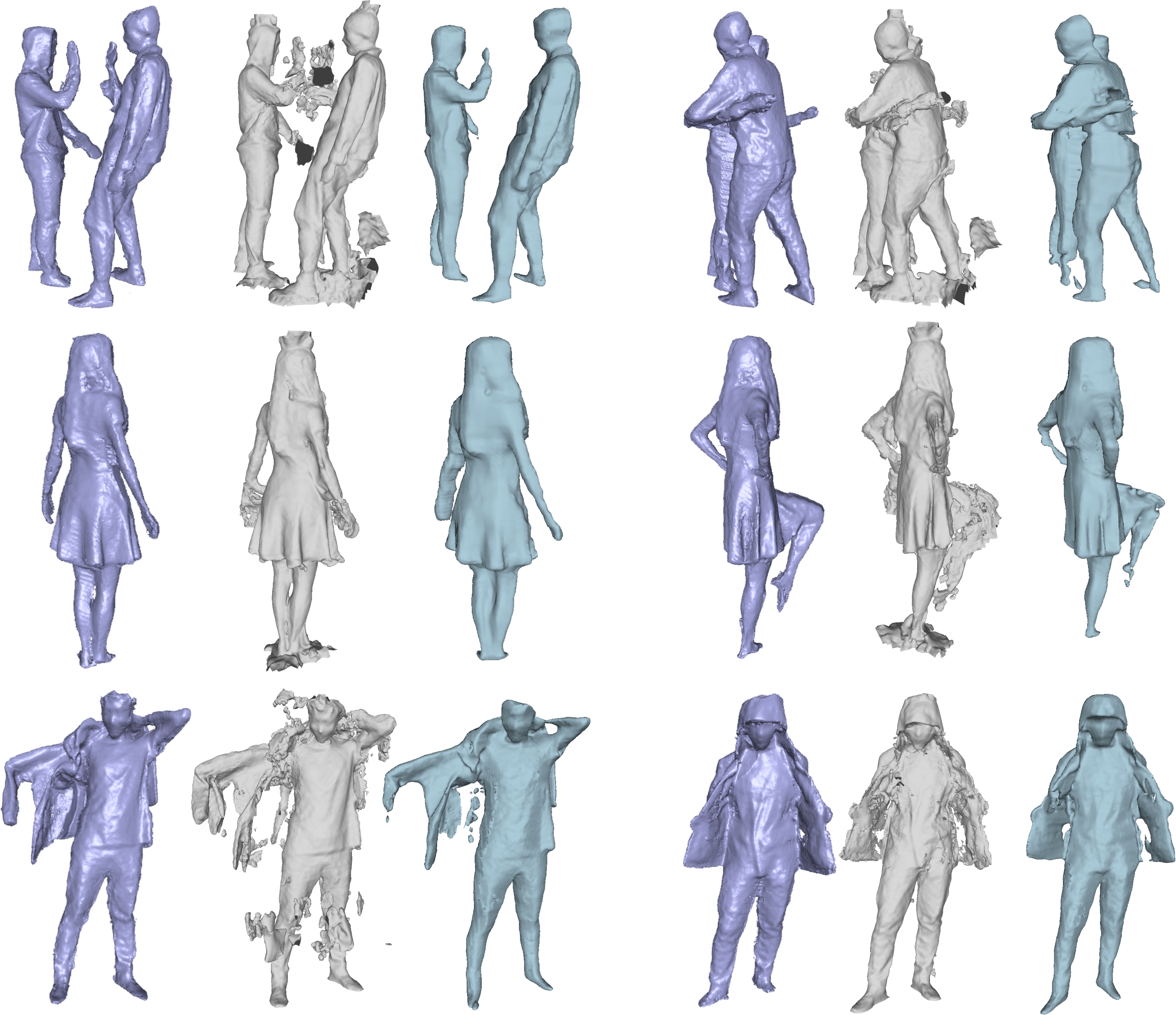}
    \caption{Qualitative comparison. For each subject, from left to right are results of our method, Motion2Fusion~\cite{Motion2Fusion} and Multi-view PIFu~\cite{saito2019pifu}, respectively. }
    \label{fig_comp}
\end{figure}

\subsection{Comparisons}

\noindent\textbf{Qualitative Comparison}
The qualitative comparison with Motion2Fusion~\cite{Motion2Fusion} and Multi-view PIFu~\cite{saito2019pifu} is shown in Fig.~\ref{fig_comp}. Given very sparse and low frame rate depth inputs from consumer RGBD sensors, Motion2Fusion generates noisy results under severe topological changing regions and fast motions due to the deteriorated non-rigid tracking performance. Moreover, the lack of depth information in Multi-view PIFu leads to over-smoothed results. 

\begin{table}
\centering
\resizebox{0.45\textwidth}{!}{
\begin{tabular}{c|c|c|c}
\hline
 Method &P2S$\times 10^{-3}${$\downarrow$} & Chamfer$\times 10^{-3}${$\downarrow$} & Normal-Consis{$\uparrow$} \\
\hline\hline
Multi-view PIFU~\cite{saito2019pifu}&$4.594$&$4.657$&$0.862$ \\
IPNet~\cite{bhatnagar2020ipnet} &$3.935$&$3.858$&$0.902$\\
\textbf{\textit{GeoNet}}&$\textbf{1.678}$&$\textbf{1.719}$&$\textbf{0.941}$\\
\hline
\end{tabular}}
\caption{Quantitative comparison of geometry reconstruction with multi-view PIFU and IPNet. 
}
\label{tab_comp}
\end{table}

\noindent\textbf{Quantitative Comparison}
We compare with $2$ state-of-the-art deep implicit surface reconstruction methods, Multi-view PIFu~\cite{saito2019pifu}(with RGB images as input) and IPNet~\cite{bhatnagar2020ipnet}(with voxelized point clouds as input). 
We retrain their networks using our training dataset and perform the evaluation on a testing dataset that contains $116$ high-quality scans with various poses, clothes, and human-object interactions. 
Tab.~\ref{tab_comp} shows the quantitative comparison results. We can see that the lack of depth information deteriorates the reconstruction accuracy of Multi-view PIFu. 
Moreover, even with multi-view depth images as input, the heavy dependency on SMPL initialization (which is difficult to get with large poses and human-object interactions) restricts the IPNet from generating highly accurate results. 
Finally, by explicitly encoding depth observations using the truncated PSDF values, the proposed \textit{GeoNet} can not only achieves accurate reconstruction results but also orders of magnitude faster than IPNet (approximately $80$ seconds for reconstruction). For a detailed description of the comparison, please refer to the supplementary material.

\subsection{Ablation Studies}

\begin{table}
\centering
\resizebox{0.45\textwidth}{!}{
\begin{tabular}{c|c|c|c}
\hline
 Method & P2S$\times 10^{-3}${$\downarrow$} & Chamfer$\times 10^{-3}${$\downarrow$} & Normal-Consis{$\uparrow$}  \\
\hline\hline
\textbf{w/o} PSDF \textbf{w.}RGBD  &$2.36$ & $2.458$ & $0.916$\\
\textbf{w/o} PSDF \textbf{w.}depth-only & $2.264$ & $2.359$ & $0.918$ \\
\textbf{w. PSDF w.depth-only} &$\textbf{1.678}$ & $\textbf{1.719}$ &$\textbf{0.941}$\\
\hline
\end{tabular}}
\caption{Ablation study on the truncated PSDF feature.}
\vspace{-15pt}
\label{tab_eval_psdf}
\end{table}

\noindent\textbf{Dynamic Sliding Fusion} As shown in Fig.~\ref{fig_eval_fusion}(a) and (b), the depth inputs in different views are not consistent with each other due to the challenging hair motion. This results in incomplete results (the orange circle). More importantly, without using dynamic sliding fusion, the result is much noisy (the red circle). By using dynamic sliding fusion, we can get more complete and noise-eliminated reconstruction results as shown in Fig.~\ref{fig_eval_fusion}(d). Please refer to the supplementary video for more clear evaluations. 

\noindent\textbf{Truncated PSDF Feature} 
The qualitative evaluation of the truncated PSDF feature is shown in Fig.\ref{fig_eval_trunc_PSDF}. 
Tab.~\ref{tab_eval_psdf} also provides quantitative evaluation results for the networks with and without using truncated PSDF values. 
We conduct two experiments with RGBD images and depth-mask images as input, respectively. We can see that without using the truncated PSDF feature, the depth-only model and RGBD model produces similar results. 
Benefiting from the truncated PSDF feature, our GeoNet achieves much accurate results, which demonstrates the effectiveness of our method. 

\begin{figure}[]
    \centering
    \includegraphics[width=0.9\linewidth]{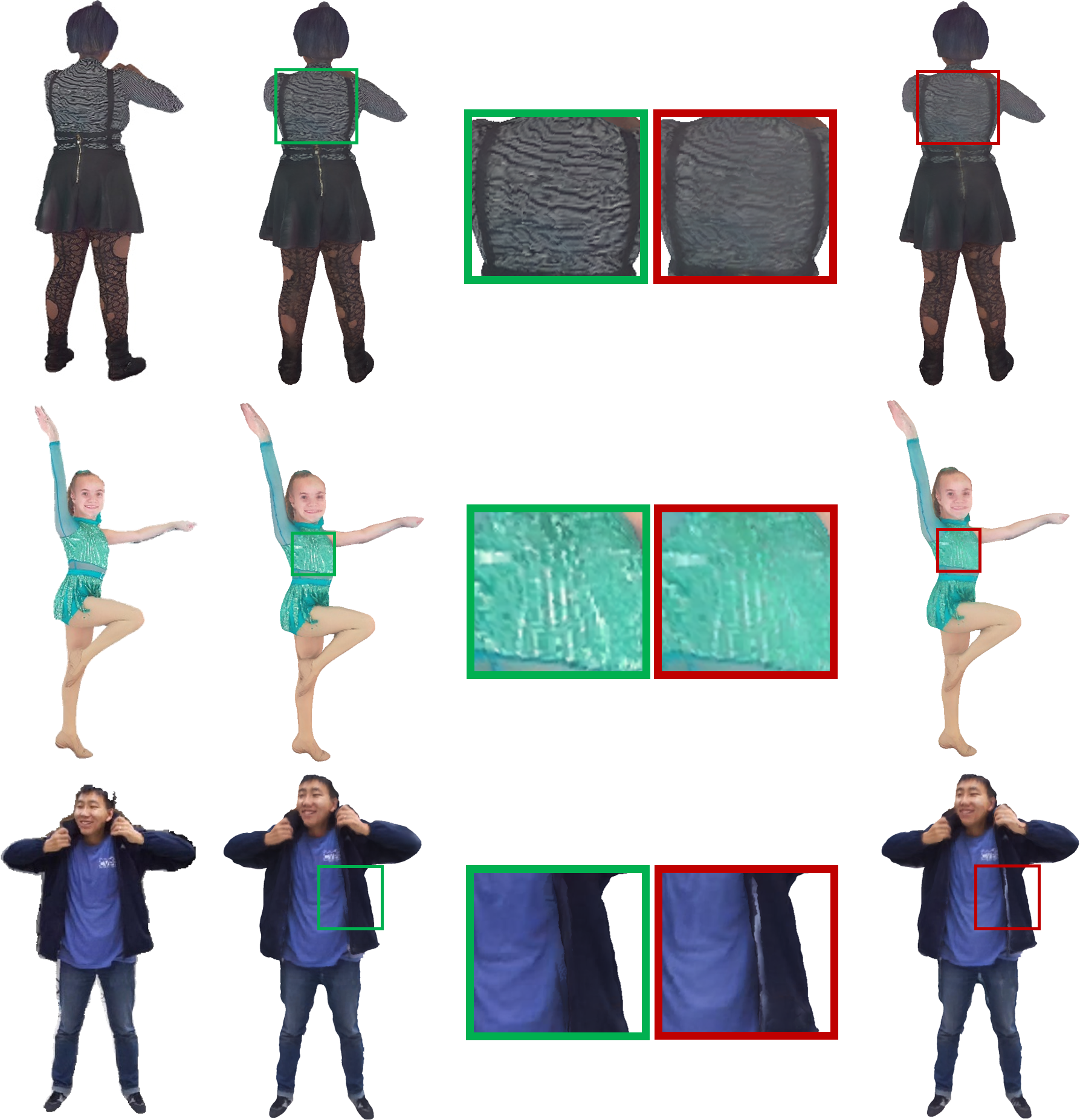}
    \caption{Evaluation of the attention mechanism in \textit{ColorNet}. From left to right are: input RGB images, texture results with (green) and without (red) attention, respectively. }
    \vspace{-20pt}
    \label{fig_eval_attn}
\end{figure}

\noindent\textbf{Attention-based Feature Aggregation}
In Fig.~\ref{fig_eval_attn}, we compare the models without using a multi-view self attention mechanism for color inference qualitatively. 
Benefiting from the multi-view self-attention mechanism, the color inference results becomes much sharper and plausible especially around observation boundaries. This is because the self-attention enables dynamic feature aggregation rather than the simple average-based feature aggregation, which enforces the MLP-based-decoder to learn how multi-view features (including geometric features and texture features) are correlated with each other in the 3D space. 

%% file: 7_conclusion.tex
\section{Conclusion}
\label{sec:conclusion}
In this paper, we propose Function4D, a real-time volumetric capture system using very sparse consumer RGBD sensors. 
By proposing dynamic sliding fusion for topology consistent volumetric fusion and detail-preserving deep implicit functions for high-quality surface reconstruction, our system achieves detailed and temporally-continuous volumetric capture even under various extremely challenging scenarios. 
We believe that such a light-weight, high-fidelity, and real-time volumetric capture system will enable many applications, especially consumer-level holographic communications, on-line education, and gaming, etc. 

\noindent\textbf{Limitations and Future Work} 
Although we can preserve the geometric details in the visible regions, generating accurate and detailed surfaces\&textures for the fully occluded regions remains challenging. This is because current deep implicit functions are mainly focus on per-frame independent reconstruction. 
Expanding deep implicit functions for using temporal observations may resolve this problem in the future. Moreover, specific materials like black hair may cause the lack of observations of depth sensors and therefore severely deteriorate current system, incorporating RGB information for geometry reconstruction may resolve this limitation and we leave this as a future work.  

\noindent \textbf{Acknowledgements} This work is supported by the National Key Research and Development Program of China No.2018YFB2100500; the NSFC No.61827805 and No.61861166002; and the China Postdoctoral Science Foundation NO.2020M670340. 
